\documentclass[lettersize,journal]{IEEEtran}
\usepackage{amsmath,amsfonts}
\usepackage{algorithmic}
\usepackage{algorithm}
\usepackage{array}
\usepackage[caption=false,font=normalsize,labelfont=sf,textfont=sf]{subfig}
\usepackage{textcomp}
\usepackage{stfloats}
\usepackage{url}
\usepackage{verbatim}
\usepackage{graphicx}
\usepackage{cite}
\usepackage{booktabs}
\usepackage{listings}
\begin{document}

\title{DSG: Dynamic 3D Scene Graph Construction \\ for Embodied Agents in Changing Indoor Environments}

\author{Ming Liao, Chao Ye, Jianing Fei, and Weiyang Lin%
\thanks{The authors are with the Research Institute of Intelligent Control and Systems,
Harbin Institute of Technology, Harbin 150001, China
(e-mail: 23b904036@stu.hit.edu.cn; yechao@hit.edu.cn;
2021112854@stu.hit.edu.cn; wylin@hit.edu.cn).}%
}


\maketitle

\begin{abstract}
In indoor environments, object positions frequently change due to human activities or embodied-agent interactions, causing previously constructed scene graphs to become inconsistent with the current scene. To address this issue, we propose DSG, a dynamic 3D scene graph construction framework that detects object changes and performs spatial relationship reasoning. First, we construct a semantic-aware 3D Gaussian scene representation and develop a dual-view rendering-based object change detection method to enable reliable scene graph node updates. Second, we propose a spatial relationship reasoning method that incorporates multi-granularity visual context, enabling a large language model to identify a richer set of inter-object spatial relationships. Furthermore, we introduce Dyn-THOR, a dynamic indoor scene graph benchmark built on the AI2-THOR simulation platform for evaluating scene graph construction in dynamic environments. Extensive experiments on Dyn-THOR, 3RScan, and real-world scenes demonstrate that DSG consistently outperforms existing methods in both object node construction and spatial relationship reasoning, significantly improving the accuracy of dynamic scene graph construction.

\end{abstract}


\section{Introduction}
\IEEEPARstart{T}{he} 3D scene graphs provide a compact representation of indoor environments by encoding the semantic attributes, geometric locations, and spatial relationships of objects. Owing to their structured representation, they have been widely adopted in a variety of downstream tasks for embodied agents, including self-localization, navigation, and robotic manipulation. However, in real-world indoor environments, object positions frequently change due to the daily activities of humans or robots. Such environmental dynamics gradually invalidate previously constructed scene graphs, resulting in inconsistencies between the scene representation and the physical environment, ultimately degrading the performance of downstream robotic tasks.

\begin{figure}[ht]
    \vspace{0mm}
	\centering
	\includegraphics[scale=0.3]{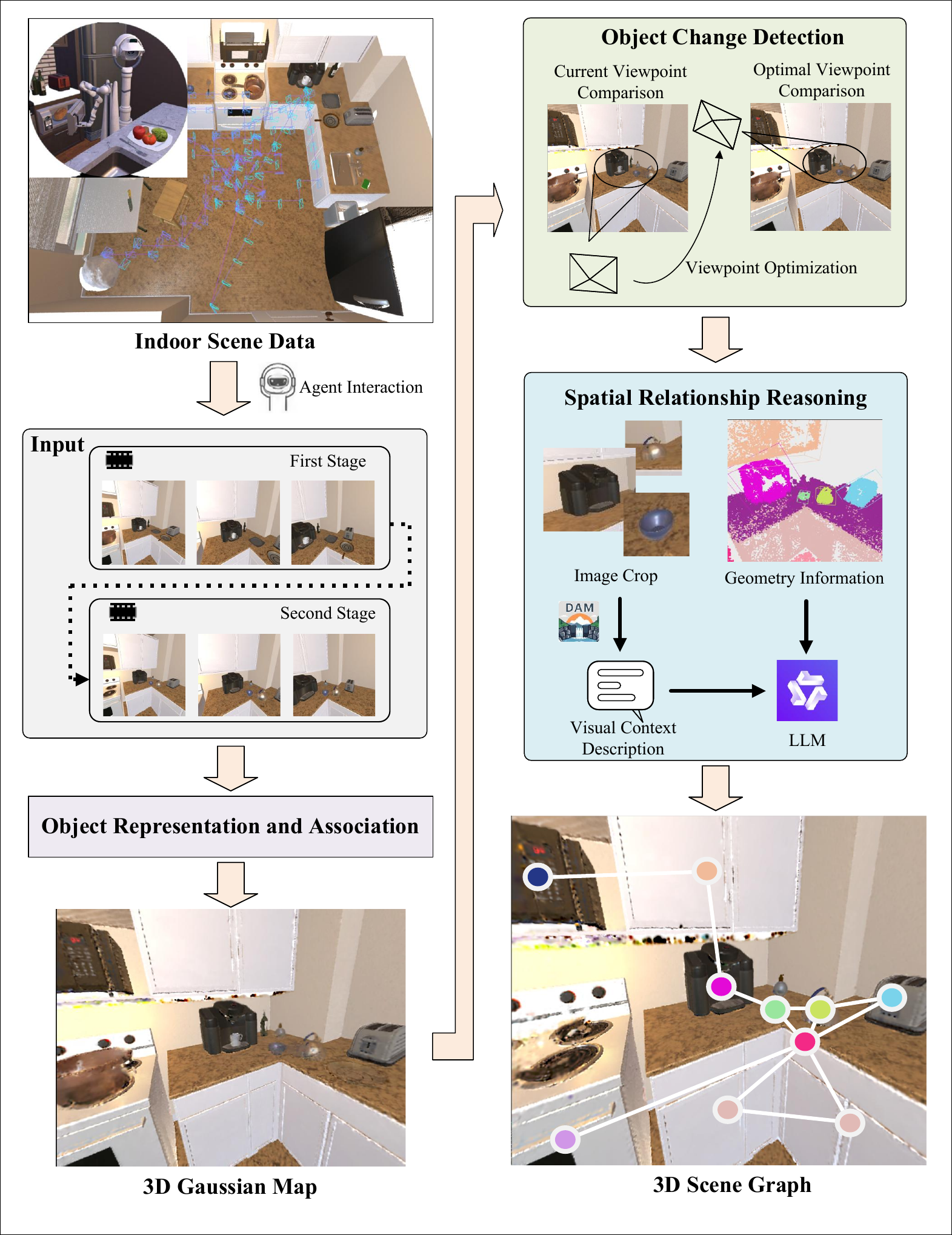}
	\vspace{-3mm}
	\caption{Overview of the proposed DSG framework. Given RGB-D sequences captured at different times, a semantic 3D Gaussian map is constructed and object changes are detected via dual-view rendering comparison. The updated object nodes are then combined with VLM descriptions and LLM reasoning to infer spatial relationships, forming the final 3D scene graph.}
	\label{overview}
	\vspace{-3mm}
\end{figure}

To achieve high-fidelity object representation, numerous scene representation methods based on 3D Gaussian Splatting (3DGS) have recently been proposed~\cite{ge2025dynamicgsg, lee2026embodiedsplat, li2025hier, li2025pg, zheng2025wildgs}. Nevertheless, most existing approaches rely heavily on offline processing and therefore struggle to update scene graphs online during robot operation. DynamicGSG~\cite{ge2025dynamicgsg} addresses this problem by rendering historical objects from the current viewpoint and comparing the rendered images with the corresponding real observations to identify missing objects. Although this rendering-based comparison is intuitive, we observe that insufficient optimization of Gaussian primitives caused by complex object geometry or material properties, as well as incomplete observations under viewpoint changes, often introduces severe rendering artifacts. Consequently, the rendered object masks deviate significantly from the corresponding image regions, leading to frequent false removal of valid objects. To overcome these limitations, we propose a dual-view rendering-based object change detection method. Instead of directly comparing rendered images before optimization, we first optimize the Gaussian primitives and then perform rendering, substantially reducing false deletions caused by rendering artifacts. Furthermore, to improve object observability, we optimize an auxiliary viewpoint by maximizing the rendered object coverage. Historical and current scenes are subsequently rendered from this optimal viewpoint for comparison, effectively alleviating missed detections caused by incomplete observations.

\IEEEpubidadjcol

After obtaining accurate 3D object representations, many scene graph construction methods~\cite{gu2024conceptgraphs, wu2021scenegraphfusion, chang2026rag, koch2024open3dsg} employ large language models (LLM) to infer spatial relationships between objects. However, these methods typically perform reasoning solely based on object labels while overlooking the rich visual context surrounding each object, resulting in inaccurate relationship prediction. To address this issue, we propose a visual-context-enhanced spatial relationship reasoning method. Specifically, we extract both object-level and surrounding-region masks from captured images and employ a vision-language model (VLM) to generate multi-granularity textual descriptions. These descriptions are further integrated with the objects’ geometric locations and 3D bounding boxes to guide an LLM in reasoning about inter-object spatial relationships, enabling richer and more complete scene graph edges.

Furthermore, existing datasets primarily evaluate scene graph construction in static environments and lack benchmarks for assessing scene graph consistency after object relocation. To fill this gap, we construct Dyn-THOR, a dynamic indoor scene graph benchmark built upon the AI2-THOR simulation platform~\cite{kolve2017ai2}. Together with a complete pipeline for data collection and object-change annotation, Dyn-THOR provides global ground-truth annotations for object relocation and establishes a quantitative benchmark for evaluating the adaptability of scene graph construction methods in dynamic environments.

In this work, we integrate Gaussian splatting-based SLAM with advanced multimodal foundation models, including GroundingDINO~\cite{liu2024grounding}, Segment Anything~\cite{kirillov2023segment}, and CLIP~\cite{radford2021learning}, to obtain high-fidelity 3D Gaussian representations of objects. We then detect object changes through rendered-image comparison across multiple viewpoints, enabling reliable identification of missing objects in dynamic environments. Subsequently, we generate spatial-context descriptions by combining object masks, surrounding visual regions, and a vision-language model, and employ a large language model to infer spatial relationships between objects accurately. Finally, we develop a complete data collection and annotation pipeline on AI2-THOR and construct the Dyn-THOR dataset, which provides precise ground-truth annotations for object changes in indoor environments. The overall framework of the proposed DSG is illustrated in Fig.~\ref{overview}.

The main contributions of this work are summarized as follows:

\begin{itemize}
\item We propose a dual-view rendering-based object change detection method that compares object renderings from both the current and optimized viewpoints, enabling reliable missing object removal and accurate scene graph updates.

\item We propose a visual-context-enhanced spatial relationship reasoning method that integrates multi-granularity visual context with 3D geometric information to improve large language model-based spatial relationship inference.

\item We introduce Dyn-THOR, a dynamic indoor scene graph benchmark built on AI2-THOR, providing object relocation scenarios and comprehensive ground-truth annotations for quantitative evaluation.

\item Extensive experiments on Dyn-THOR, 3RScan, and real-world scenes demonstrate that the proposed method consistently outperforms existing approaches in object node construction and spatial relationship reasoning.
\end{itemize}

\begin{figure*}[ht]
    \vspace*{0mm}
	\centering
	\includegraphics[scale=0.26]{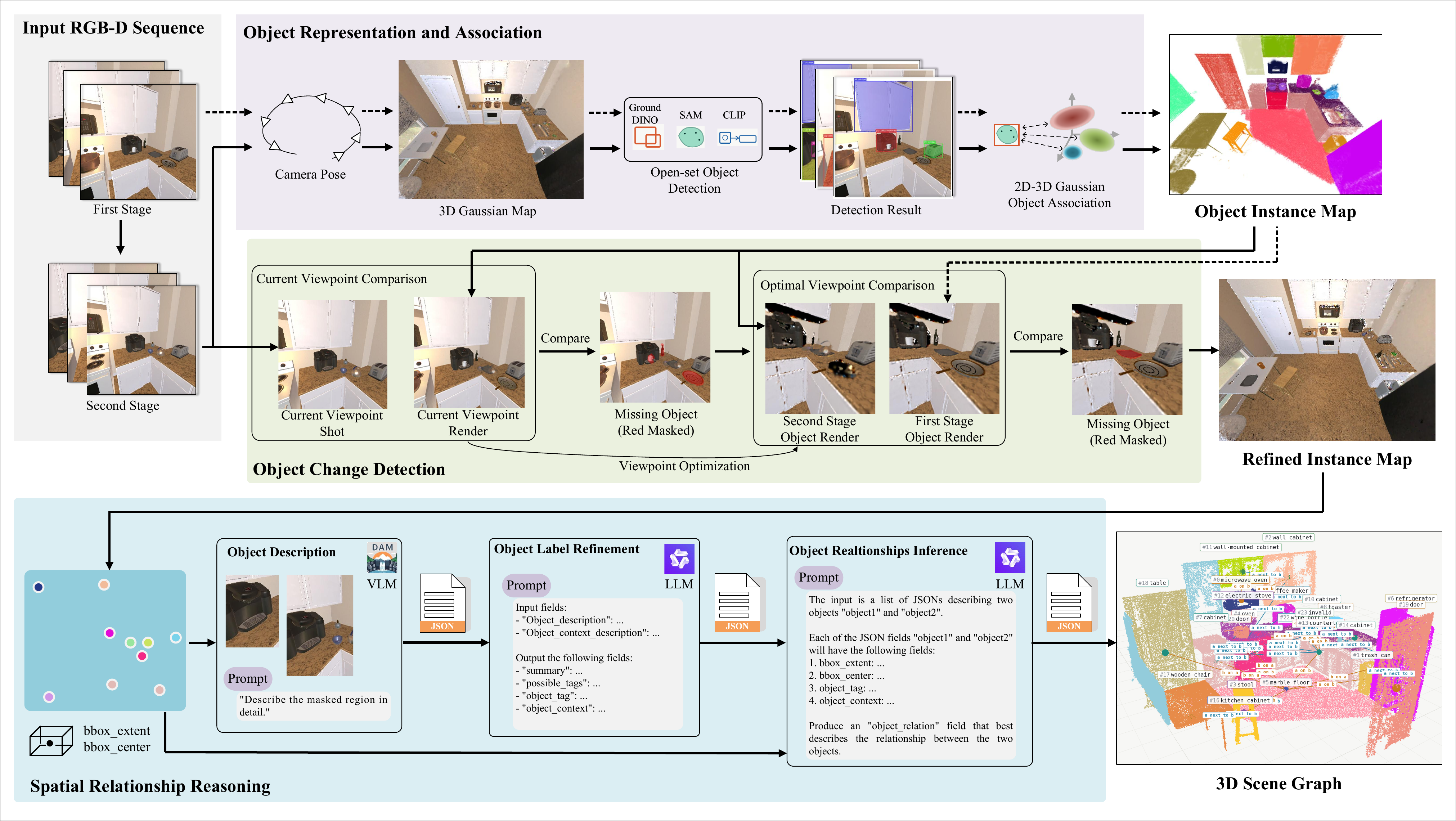}
	\vspace{-2mm}
	\caption{Pipeline of the proposed framework. Input RGB-D sequences are processed by open-vocabulary perception models (Grounding-DINO, SAM, and CLIP) and associated with 3D Gaussian primitives to construct a semantic Gaussian map. Object changes are then detected through progressive rendering comparison from the current viewpoint and an optimized viewpoint, yielding an updated object set. For each object, visual descriptions are generated from image observations and refined into structured semantic representations. Finally, the object semantics, visual context, and geometric attributes are jointly fed into a large language model to infer inter-object spatial relationships, producing the final 3D scene graph.}
	\label{pipline}
	\vspace{-3mm}
\end{figure*}

\section{Related Work}

\subsection{3D Object Representation}

Efficient and high-fidelity object representation forms the foundation of perception and planning for embodied agents. Early approaches primarily represented scenes using voxels, point clouds, or meshes. For example, Kimera~\cite{rosinol2021kimera} and Hydra~\cite{hughes2022hydra} construct metric maps based on Euclidean Signed Distance Functions (ESDFs), but such representations inevitably involve trade-offs among memory consumption, reconstruction quality, and object representation fidelity. 

More recently, 3DGS has emerged as a powerful scene representation owing to its high rendering quality and real-time rendering capability. Representative methods such as SplaTAM~\cite{keetha2024splatam} and CG-SLAM~\cite{hu2024cg} exploit the differentiable rendering property of 3DGS to achieve high-fidelity online mapping, while LangSplat~\cite{qin2024langsplat} and GaussianGrouping~\cite{ye2024gaussian} further incorporate vision-language features into Gaussian primitives to enhance semantic scene understanding. DQO-MAP~\cite{li2025dqo} combines dual quadrics and 3DGS for real-time object pose estimation and reconstruction. CSGrasp~\cite{ye2025csgrasp} transfers category-level semantic and grasp priors through point-wise alignment. However, most existing methods still rely heavily on offline processing and therefore struggle to update object representations online during robot operation. DynamicGSG~\cite{ge2025dynamicgsg} detects missing objects by comparing rendered masks with observations, but its single-threshold criterion is sensitive to viewpoint, appearance, and Gaussian optimization artifacts, causing erroneous removals. These limitations motivate the development of a more robust object change detection strategy for dynamic scene graph construction.

\subsection{3D Scene Graph Construction}

A 3D scene graph represents an environment as a graph consisting of object nodes and relationship edges, providing a compact and structured representation for embodied perception and decision making.

Early scene graph construction methods~\cite{rosinol20203d, kim20193} primarily generated object-centric scene graphs from RGB-D observations or reconstructed 3D scenes by assigning semantic and geometric attributes to object nodes. Qi et al.~\cite{qi2023instance} formulate instance-incremental 3D scene graph generation as a conditional generation problem, autoregressively adding novel object instances and their relationships to point-cloud scenes via normalizing flows. Subsequent works, such as S-Graphs~\cite{bavle2023s} and HOV-SG~\cite{werby2024hierarchical}, further introduced hierarchical representations by incorporating room- and building-level nodes, enabling large-scale scene understanding and navigation. Zhang et al.~\cite{zhang2026long} jointly model objects, planes, and lines for robust graph matching and map alignment. Feng et al.~\cite{feng2025history} incrementally construct a consistent 3D semantic scene graph from RGB-D sequences by exploiting global and local historical information for temporal relationship reasoning.

Recent studies have increasingly employed VLM and LLM to improve spatial relationship reasoning. ConceptGraphs~\cite{gu2024conceptgraphs} infer relations from geometric overlap and semantic labels, while Open3DSG~\cite{koch2024open3dsg}  further exploit vision-language features or textual scene descriptions to enhance reasoning capability. Hou et al.~\cite{hou2026spatial} leverage a fine-tuned VLM to infer object support relations and construct hierarchical visual and symbolic graphs, progressively integrating contextualized visual features with textual knowledge for 3D relationship prediction. KeySG~\cite{werby2025keysg} enriches object semantics through key-frame descriptions, improving semantic completeness for relationship prediction. MM-SGG~\cite{lv2023multimodality} uses a heat modality to encode spatial cues for visual relationship prediction.

Despite these advances, most existing approaches rely primarily on object labels, geometric cues, or abstract scene-level context, while insufficiently exploiting observation-specific visual context between object pairs for general spatial relationship reasoning, limiting their ability to distinguish ambiguous relationships in complex indoor environments. Furthermore, they generally assume static environments and therefore lack mechanisms for maintaining scene graph consistency after object relocation, motivating more robust scene graph construction methods for dynamic environments.

Dynamic indoor environments introduce additional challenges for scene graph construction because object relocation continuously changes the scene topology. L-DOR~\cite{zhang2026long} models spatio-temporal object distributions for long-term dynamic relocalization. Although recent methods have begun to consider dynamic scene understanding~\cite{li2025pg, zheng2025wildgs, schmid2024khronos}, online scene graph maintenance remains largely underexplored. Moreover, existing benchmarks are primarily designed for static environments and provide limited support for evaluating scene graph construction after object relocation. The lack of quantitative benchmarks with object change annotations makes it difficult to comprehensively assess the robustness of dynamic scene graph construction methods.

\section{Method}

The proposed framework named DSG, aims to construct a globally consistent 3D scene graph in dynamic indoor environments where object locations may change over time. As illustrated in Fig.~\ref{pipline}, the framework consists of three components. First, a semantic-aware Gaussian map is constructed to provid a high-fidelity 3D representation for each object (Section~\ref{subsec:object_rep}). Second, a dual-view rendering-based object change detection module identifies missing objects by comparing rendered observations from both the current and an optimized viewpoint, yielding an updated object set (Section~\ref{subsec:object_change}). Finally, multi-granularity visual-language descriptions are combined with the LLM to infer spatial relationships between objects and construct scene graph edges (Section~\ref{subsec:spatial_relation}).

\subsection{Object Representation and Association}
\label{subsec:object_rep}

Given a posed RGB-D sequence, DSG incrementally constructs a semantic-aware Gaussian map in which each object is represented by an independent set of Gaussian primitives.

For each input frame $I_t$ at time $t$, the open-vocabulary detector GroundingDINO first predicts an object bounding box $b_{t,i}$ for object $o_i$. Segment Anything (SAM) is then employed to generate the corresponding instance mask $m_{t,i}$. To encode semantic information, the masked object region is fed into CLIP to extract an instance-level visual feature $f_{t,i}$. Let $g_j$ denote an existing Gaussian object in the global map with semantic feature $f_{g_j}$. Projecting $g_j$ into the current camera view produces a rendered object mask $m_{t,j}^{\mathrm{render}}$. The geometric similarity between the rendered mask and the detected mask is computed as
\begin{equation}
s_{\mathrm{geometry}}(i,j)=
\frac{m_{t,i}\cap m_{t,j}^{\mathrm{render}}}
{m_{t,i}\cup m_{t,j}^{\mathrm{render}}}
\end{equation}
while the semantic similarity is defined as
\begin{equation}
s_{\mathrm{semantic}}(i,j)
=
\frac{f_{t,i}^{T}f_{g_j}}{2}+0.5
\end{equation}

The overall matching score is computed by
\begin{equation}
s(i,j)
=
s_{\mathrm{geometry}}(i,j)
+
s_{\mathrm{semantic}}(i,j)
\end{equation}
and object association is performed greedily according to the overall similarity score. If a successful match is found, the global semantic feature is updated by a weighted average:
\begin{equation}
f_{g_j}
=
\frac{n_{g_j}f_{g_j}+f_{t,i}}
{n_{g_j}+1}
\end{equation}
where $n_{g_j}$ denotes the number of observations associated with Gaussian object $g_j$. Otherwise, if no existing object satisfies the matching criterion, the detection is initialized as a new Gaussian object. Each object is represented by a set of Gaussian primitives sharing the same object identifier. The Gaussian parameters, including position, color, scale, rotation, and opacity, are optimized through differentiable rendering using the captured RGB images, resulting in a high-fidelity object representation.

\subsection{Object Change Detection}
\label{subsec:object_change}

Given a historical Gaussian map and the current RGB observation, the objective is to determine whether an object remains stable, is missing from its previous location, or should be initialized as a newly observed object. We propose a dual-view rendering-based change detection method consisting of two stages: comparison under the current viewpoint, and verification from an optimized viewpoint.

\textbf{Current Viewpoint Comparison:} Unlike DynamicGSG, which compares renderings before optimizing the map, we first optimize the Gaussian primitives using the rendering loss and then perform object difference detection based on the optimized renderings. For an object $g_j$ visible within the current view frustum, the optimized map rendered into the current viewpoint, $\hat{I}_t$, yields the visible object mask $\hat{m}_{t,j}$, which is compared against the corresponding RGB region $I_t(\hat{m}_{t,j})$ via channel-wise SSIM:
\begin{equation}
\lambda_j = \frac{1}{3}\sum_{c \in \{R,G,B\}} \mathrm{SSIM}\!\left(\hat{m}_{t,j}^{(c)},\, I_t(\hat{m}_{t,j})^{(c)}\right)
\end{equation}
Based on $\lambda_j$, each object is classified as
\begin{equation}
\mathrm{State}(g_j) =
\begin{cases}
\text{Stable} & \lambda_j \ge \tau_{high} \\
\text{Uncertain} & \tau_{low} \le \lambda_j \le \tau_{high} \\
\text{Missing} & \lambda_j \le \tau_{low}
\end{cases}
\end{equation}
\begin{figure}[ht]
    \vspace*{0mm}
	\centering
	\includegraphics[scale=0.6]{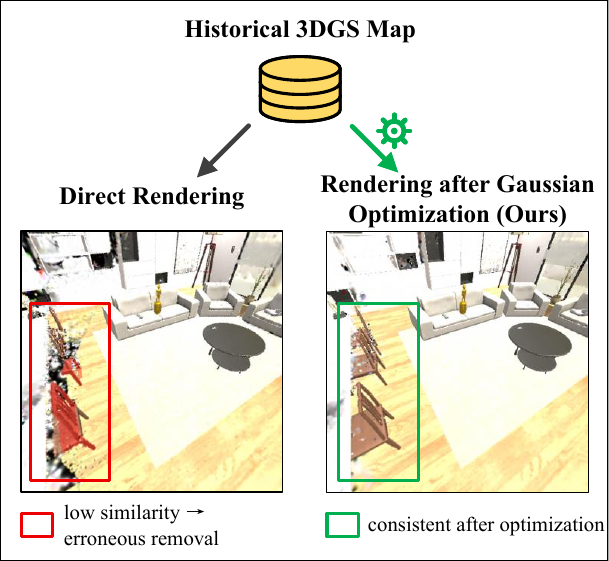}
	\vspace{-3mm}
	\caption{Comparison between direct rendering and rendering after Gaussian optimization. Due to incomplete observations and rendering artifacts, direct rendering may produce inaccurate object appearances, leading to erroneous removal of valid objects. By optimizing Gaussian primitives before rendering, the proposed method significantly reduces rendering artifacts and improves object consistency.}
	\label{ai2thor_render_visual}
	\vspace{-3mm}
\end{figure}
where $\tau_{high}$ and $\tau_{low}$ are the high- and low-confidence thresholds. Objects classified as uncertain are further verified from an optimized viewpoint.

Rendering from optimized rather than raw Gaussian primitives reduces artifacts caused by limited viewpoints or complex object materials, leading to more reliable object comparison, as shown in Fig.~\ref{ai2thor_render_visual}. At the same time, even after an object is missing from its original position, its Gaussian primitives are not immediately removed during optimization and still retain enough information to support the decision made at this stage.

\begin{figure}[ht]
    \vspace*{0mm}
	\centering
	\includegraphics[scale=0.5]{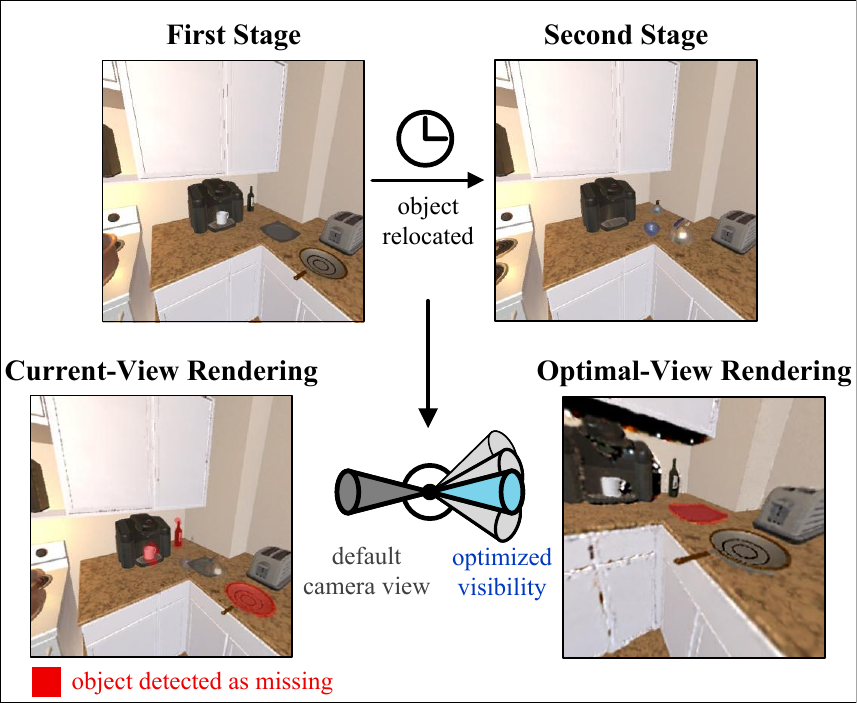}
	\vspace{-3mm}
	\caption{The figure illustrates the complementary roles of current-view and optimal-view rendering for object change detection. While current-view rendering may fail to detect missing objects due to unfavorable viewing conditions, optimal-view rendering provides more complete observations, enabling accurate detection of object changes.}
	\label{ai2thor_optimalview_visual}
	\vspace{-3mm}
\end{figure}

\textbf{Optimal Viewpoint Comparison:} The current-view comparison may become unreliable when an object is only partially observed. To further verify uncertain objects, we estimate an optimized viewpoint that maximizes object visibility and compare the renderings of the historical and current Gaussian maps from this viewpoint, as shown in Fig.~\ref{ai2thor_optimalview_visual}. Specifically, starting from the current camera pose, a virtual camera pose $(\mathbf{R}^*,\mathbf{t}^*)$ is optimized as follows:
\begin{equation}
\mathbf{R}^*,\mathbf{t}^*
=
\operatorname*{arg\,min}_{\mathbf{R},\mathbf{t}}
\left(
L_{frustum}(\mathbf{R},\mathbf{t})
+
\alpha
L_{vis}(\mathbf{R},\mathbf{t})
\right)
\end{equation}
where $L_{frustum}$ encourages complete object observation, $L_{vis}$ constrains the observation distance, and $\alpha$ is a weighting coefficient.

The frustum loss $L_{frustum}$ encourages all uncertain objects to remain within the camera field of view while placing their projections close to the image center. Let
$P_i=\{\mathbf{p}_k\in\mathbb{R}^3\}$ denote the Gaussian mean points of uncertain object $i$. Under the virtual camera pose $(\mathbf{R},\mathbf{t})$, each point is transformed into the camera coordinate system as
\begin{equation}
\mathbf{p}_k^{c}=\mathbf{R}\mathbf{p}_k+\mathbf{t}
\end{equation}
where
$\mathbf{p}_k^{c}=(p_k^x,p_k^y,p_k^z)^T$. The projected image coordinates are
\begin{equation}
u_k=f_x\frac{p_k^x}{p_k^z}+c_x,\qquad
v_k=f_y\frac{p_k^y}{p_k^z}+c_y
\end{equation}
where $f_x,f_y$ are the focal lengths and $c_x,c_y$ denote the principal point. The image coordinates are normalized as
\begin{equation}
\tilde{u}_k=\frac{2u_k}{W}-1,\qquad
\tilde{v}_k=\frac{2v_k}{H}-1
\end{equation}
where $W$ and $H$ are the image width and height. A differentiable visibility score is then defined as
\begin{equation}
s_k(\mathbf{R},\mathbf{t})=\sigma\!\left(\frac{1-\tilde{u}_k^2}{\delta}\right)\cdot\sigma\!\left(\frac{1-\tilde{v}_k^2}{\delta}\right)
\end{equation}
where $\delta$ is a temperature parameter and $\sigma(\cdot)$ denotes the sigmoid function.

The frustum loss is defined as follows:
\begin{equation} L_{frustum}(\mathbf{R},\mathbf{t})
=\sum_i\left(1-\frac{1}{|P_i|}\sum_{k\in P_i}s_k(\mathbf{R},\mathbf{t})\right)^2
\end{equation}
minimizing $L_{frustum}$ encourages the projected Gaussian points of all uncertain objects to remain close to the image center, thereby maximizing observation coverage.

However, optimizing $L_{frustum}$ alone may place the virtual viewpoint excessively far from the target objects. To avoid this degenerate solution, a visibility regularization term is introduced to maintain an appropriate observation distance. During differentiable rendering, Gaussian primitives belonging to uncertain objects are rendered in white, while all remaining primitives are rendered in black, producing a semantic rendering
$\hat{I}(\mathbf{R},\mathbf{t})\in[0,1]^{H\times W}$.
The visibility loss is defined as follows:
\begin{equation}
L_{vis}(\mathbf{R},\mathbf{t})=-\frac{1}{HW}\sum_{h,w}\hat{I}(\mathbf{R},\mathbf{t})(h,w)
\end{equation}
minimizing $L_{vis}$ encourages the target objects to occupy a larger image region, preventing the optimized viewpoint from drifting excessively far away.

After obtaining the optimized camera pose $(\mathbf{R}^*,\mathbf{t}^*)$, the historical and current Gaussian maps are rendered into this viewpoint, producing $\hat{I}_{before}$ and $\hat{I}_{after}$, respectively. For each uncertain object, the rendered object mask $\hat{m}_j$ is extracted and the structural similarity is computed as follows:
\begin{equation}
\lambda_j^*=\frac{1}{3}\sum_{c\in\{R,G,B\}}\mathrm{SSIM}\!\left(\hat{I}_{before}(\hat{m}_j),\hat{I}_{after}(\hat{m}_j)\right)
\end{equation}

If $\lambda_j^*<\tau_{low}$, the object is classified as Missing; otherwise, it is considered Stable. Newly observed objects are initialized through the object association procedure described in Section~3.1 and are classified as Appeared. Consequently, every object is assigned to one of three states: Stable, Appeared or Missing.

\subsection{Spatial Relationship Reasoning}
\label{subsec:spatial_relation}

After object change detection, the updated Gaussian objects constitute the nodes of the scene graph. The remaining task is to infer the spatial relationships between objects and construct the graph edges. Existing methods typically rely on object labels and geometric information for relationship reasoning, limiting their ability to capture the rich visual context surrounding objects. To address this limitation, we propose a spatial relationship reasoning method that combines multi-granularity visual descriptions with the LLM.

\textbf{Visual Descriptions:}
For each Gaussian object, we first select the observation in which the object occupies the largest image region and extract the corresponding instance mask from the captured RGB image. Two complementary textual descriptions are then generated using the Describe Anything Model (DAM)~\cite{lian2025describe}. The first is an object-level description, which focuses on the instance mask and captures the object's appearance, color, material, and other intrinsic attributes. The second is a context-level description. Specifically, the object mask is expanded by a fixed factor of 2.5, and the enlarged image region is provided to DAM, allowing the generated description to include neighboring objects and the surrounding spatial context. The two descriptions are subsequently fed into LLM together with carefully designed few-shot prompts. The model outputs a structured representation containing the object category and its visual context, which serves as the textual input for subsequent spatial relationship inference.

\textbf{Spatial Relationship Inference:}
To reduce unnecessary reasoning, candidate object pairs are first generated according to their geometric proximity. For every pair of objects, the 3D bounding-box IoU is computed. Object pairs with zero IoU are discarded directly, while pairs with nonzero IoU are further verified using the point-cloud overlap ratio computed through FAISS nearest-neighbor search. Pairs with insufficient overlap are removed, yielding the final candidate edge set. For each candidate pair, the 3D geometric information, semantic labels, and multi-granularity visual descriptions are jointly provided to the LLM for spatial relationship inference. The LLM combines geometric cues, including relative position, height difference, and spatial overlap, with the visual context extracted from the captured images to infer the relationship between the two objects. The inferred relationship label is then assigned as an edge in the scene graph. Together with the updated Gaussian object nodes, the complete 3D scene graph is finally represented as $G=(V,E)$, where $V$ and $E$ denote the sets of object nodes and spatial relationship edges.

\section{Experiments}

\subsection{Datasets}

Experiments are conducted on three datasets: the proposed Dyn-THOR benchmark, the real-world 3RScan dataset~\cite{wald2019rio}, and a self-collected real-world dataset.

\textbf{Dyn-THOR:}
Dyn-THOR is a dynamic indoor scene benchmark built upon the AI2-THOR simulation platform~\cite{kolve2017ai2}. It covers four common indoor scene categories, including kitchens, living rooms, bedrooms, and bathrooms, and contains 20 sequences with a total of 3326 keyframes. Each keyframe provides synchronized RGB images, depth maps, and ground-truth camera poses. Each sequence consists of two stages. In the first stage, all objects remain in their original positions. In the second stage, up to ten movable objects are randomly relocated to simulate object changes caused by human or agent activities. Both stages share identical camera trajectories, enabling direct comparison before and after object relocation. In addition, each sequence provides object-level ground-truth annotations for three object states (Stable, Appeared, and Missing), together with ground-truth 3D bounding boxes and annotated spatial relationships. These annotations enable quantitative evaluation of dynamic scene graph construction.

\textbf{3RScan:}
3RScan is a large-scale real-world RGB-D dataset consisting of 1,482 indoor scene scans, where each scene contains a reference scan together with one or more rescans captured at different time instances. The dataset provides RGB-D sequences, ground-truth camera poses, and object instance annotations, making it suitable for evaluating scene graph construction under object relocation. We select five reference--rescan scene pairs for evaluation, comprising a total of 1866 keyframes.

\textbf{Real-world dataset:}
To further validate the proposed method in real environments, we collect an RGB-D dataset using an Intel RealSense D455 camera. Both RGB and depth images are captured at a resolution of $640 \times 480$, while camera poses are estimated using VINS-Fusion~\cite{qin2025general}. The resulting dataset contains a total of 831 frames covering diverse indoor objects and layouts.

\subsection{Experimental Setup}

\textbf{Evaluation metrics:}
For node-level evaluation, we first compute the 3D IoU between the predicted and ground-truth object bounding boxes and perform one-to-one object matching using the Hungarian algorithm. The semantic consistency of each matched object pair is further verified using CLIP similarity between their semantic labels. Based on the remaining valid matches, we report \textbf{Precision}, \textbf{Recall}, and \textbf{F1 score}. To evaluate the capability of detecting object relocation, we further introduce the \textbf{Missing Residual Rate (MRR)}, which measures the percentage of ground-truth missing objects that remain in the reconstructed map; a lower MRR indicates better removal of missing objects.

\begin{table}[h]
\centering
\caption{Node-level evaluation and ablation study on the Dyn-THOR dataset. P: Precision, R: Recall, F1: F1 Score, MRR: Missing Residual Rate.}
\label{ai2thor_nodes_table}
\begin{tabular*}{\columnwidth}{@{\extracolsep{\fill}}lcccc}
\toprule
Method & P$\uparrow$ & R$\uparrow$ & F1$\uparrow$ & MRR$\downarrow$ \\
\midrule
ConceptGraphs~\cite{gu2024conceptgraphs} & 33.3 & 26.0 & 28.3 & 14.9 \\
DynamicGSG~\cite{ge2025dynamicgsg}       & 44.8 & 29.5 & 35.2 & 3.0  \\
\midrule
Ours w/o Remove Object          & 49.5 & \textbf{34.7} & 40.4 & 9.3 \\
Ours w/ Single-view Remove  & \textbf{52.8} & 33.7 & \textbf{40.7} & 2.6 \\
Ours w/ Dual-view Remove (Full)   & 51.0 & 33.9 & 40.3 & \textbf{2.1} \\
\bottomrule
\end{tabular*}
\end{table}

\begin{table}[h]
\centering
\caption{Node-level evaluation results on the 3RScan dataset.}
\label{3rscan_nodes_table}
\begin{tabular*}{\columnwidth}{@{\extracolsep{\fill}}lccc}
\toprule
Method & Precision$\uparrow$ & Recall$\uparrow$ & F1 Score$\uparrow$ \\
\midrule
DynamicGSG~\cite{ge2025dynamicgsg} & 15.3 & 10.2 & 11.7 \\
Ours                                & \textbf{17.5} & \textbf{34.1} & \textbf{22.8} \\
\bottomrule
\end{tabular*}
\end{table}

For edge-level evaluation, we compare the spatial relationships predicted by different methods with the ground truth. An edge is considered \textbf{Evaluable} only when both endpoint objects are successfully matched. We first measure the number of evaluable edges to assess graph completeness. A predicted edge is regarded as a \textbf{Coarse Matches} if a relationship exists between the corresponding object pair. It is further counted as a \textbf{Fine Matches} only when the subject, predicate, and object are all consistent with the ground truth.

\textbf{Implementation details:}
Before evaluation, the object point clouds generated by all methods are voxel-downsampled and denoised using DBSCAN. The voxel size is set to 0.01\,m, the DBSCAN radius parameter is set to 0.1\,m, and the minimum number of points is set to 10. The confidence thresholds in the object change detection stage are set to $\tau_{high}=0.9$ and $\tau_{low}=0.3$. For node evaluation, the IoU threshold and CLIP semantic-similarity threshold are set to 0.3 and 0.75, respectively. For edge evaluation, all methods use Qwen3-VL-Plus for spatial relationship reasoning to ensure a fair comparison. All experiments are conducted on a workstation equipped with an NVIDIA RTX 3090 GPU.

\begin{figure*}[ht]
    \vspace*{0mm}
	\centering
	\includegraphics[scale=0.93]{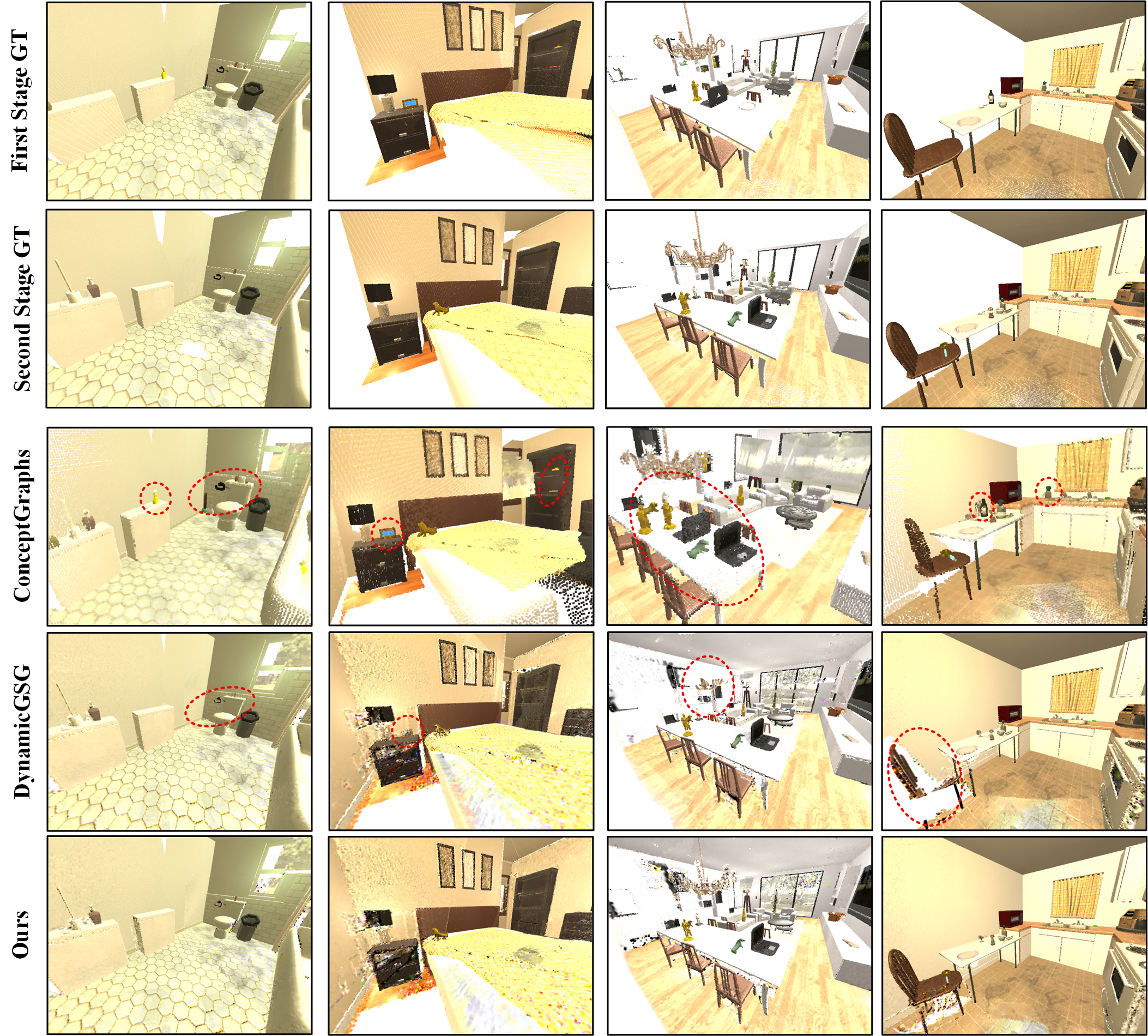}
	\vspace{-3mm}
	\caption{Qualitative comparison of object on the Dyn-THOR dataset. Red dashed circles highlight incorrect object reconstructions. ConceptGraphs retains obsolete objects after scene changes, DynamicGSG suffers from missed detections and erroneous deletions, while the proposed method produces the most accurate object representations.}
	\label{ai2thor_nodes_visual}
	\vspace{-3mm}
\end{figure*}

\begin{figure}[ht]
    \vspace*{-3mm}
	\centering
	\includegraphics[scale=0.7]{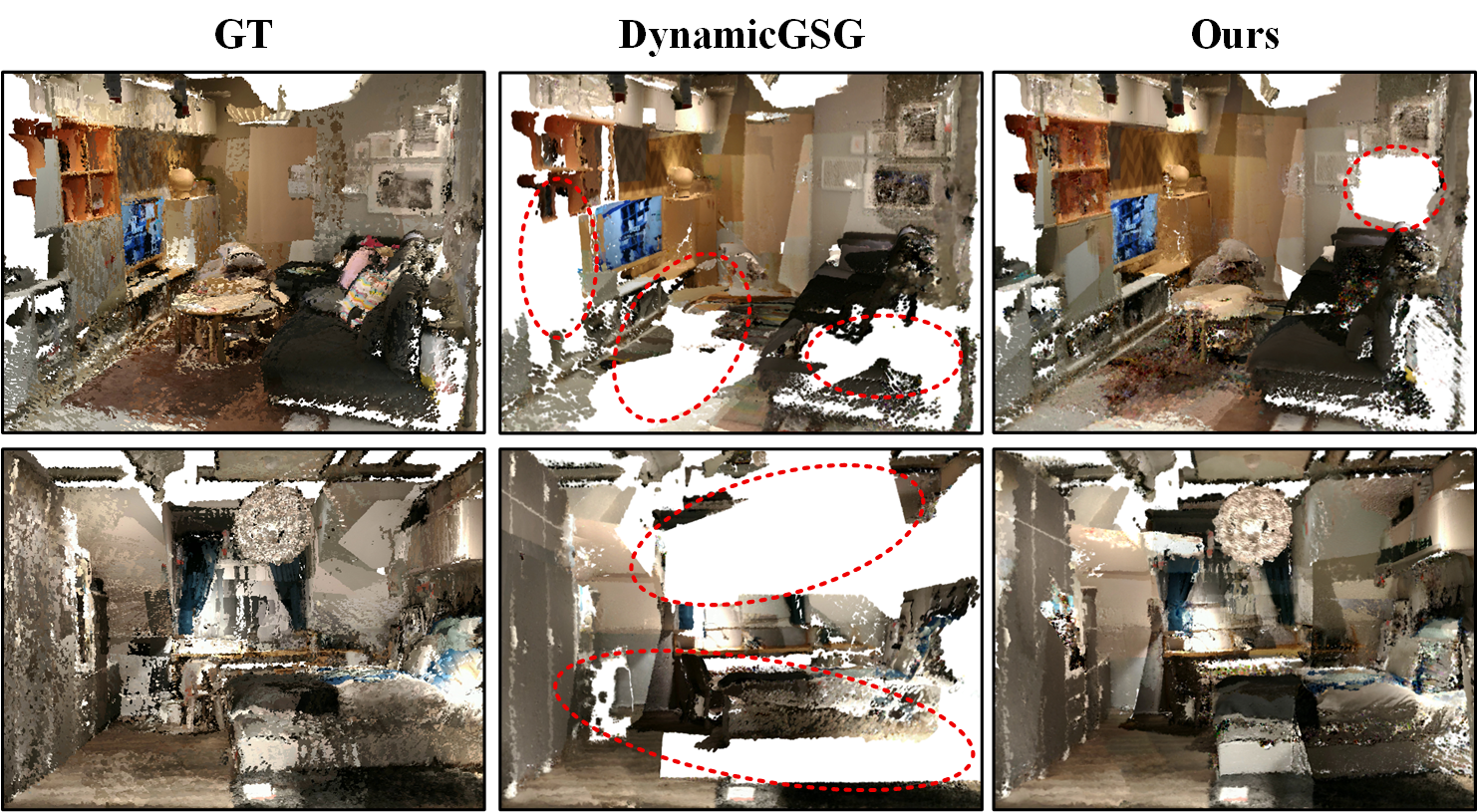}
	\vspace{-5mm}
	\caption{Qualitative comparison of object on the 3RScan dataset. Red dashed circles indicate objects that are incorrectly removed. DynamicGSG exhibits frequent erroneous deletions under challenging real-world conditions, whereas the proposed method preserves more complete object representations.}
	\label{3rscan_nodes_visual}
	\vspace{-5mm}
\end{figure}

\subsection{Node-Level Evaluation}

We first evaluate the quality of scene graph nodes after object relocation. The experiments are conducted on the Dyn-THOR and 3RScan datasets, comparing the proposed method with ConceptGraphs~\cite{gu2024conceptgraphs} and DynamicGSG~\cite{ge2025dynamicgsg}.

Tables~\ref{ai2thor_nodes_table} and ~\ref{3rscan_nodes_table} show that the proposed method consistently outperforms the existing baselines in Precision, Recall, and F1 score on both Dyn-THOR and 3RScan. In particular, the full dual-view variant achieves the lowest MRR of 2.1\% on Dyn-THOR while retaining Precision, Recall, and F1 scores comparable to those of the single-view variant. The proposed dual-view rendering strategy reliably removes Gaussian primitives belonging to missing objects, preventing residual primitives from forming ghost nodes that would otherwise occupy the matching slots of ground-truth objects during Hungarian matching. Moreover, removing these residual primitives before object association prevents newly appeared objects from being incorrectly merged with historical objects, thereby reducing ghost nodes and preserving more reliable object associations. In contrast, DynamicGSG relies on a single-view rendering comparison, which is more susceptible to rendering artifacts and incomplete observations, leading to both erroneous object deletion and residual missing objects. Since ConceptGraphs assumes a static environment and does not update object nodes online, missing objects always remain in the reconstructed map after object relocation, resulting in the poorest node-level performance.

The proposed method also achieves the lowest MRR, indicating a stronger capability to identify and remove missing objects. This improvement mainly benefits from two aspects. First, object comparison is performed after Gaussian optimization, which significantly reduces rendering artifacts caused by limited viewpoints and challenging object materials. Second, an optimized virtual viewpoint is introduced to maximize object visibility, enabling uncertain objects to be verified from a more informative viewpoint when the current observation is incomplete. As a result, missing objects can be detected more reliably while avoiding erroneous deletions caused by insufficient observations. By contrast, DynamicGSG performs object comparison only from the current viewpoint before optimization, making it more sensitive to rendering noise and viewpoint variations.

Furthermore, Table~\ref{ai2thor_nodes_table} presents an ablation study on the proposed object-culling strategy, removing object culling entirely causes MRR to rise sharply from 2.1 to 9.3, showing that the underlying Gaussian pruning operations alone can only clear a limited portion of residual Gaussians. The accompanying drop in Precision and Recall further indicates that relying solely on this Gaussian-level pruning is insufficient: objects that are not properly removed persist as ambiguous candidates during matching, which in turn lowers detection precision. Compared with the single-view variant, the full dual-view strategy combines comparisons from the current and optimized viewpoints, reducing the MRR from 2.6\% to 2.1\%. Meanwhile, Recall remains comparable, increasing slightly from 33.7\% to 33.9\%, whereas Precision and F1 decrease modestly from 52.8\% to 51.0\% and from 40.7\% to 40.3\%, respectively. These results indicate a trade-off between more reliable missing-object removal and the preservation of all valid objects. The slight decrease in Precision suggests that additional viewpoint verification may introduce a small number of false object removals in ambiguous cases. Nevertheless, this trade-off aligns with the objective of maintaining scene graph consistency after object relocation, where reliably eliminating obsolete objects is crucial while preserving comparable overall node-construction quality.

\begin{figure*}[ht]
    \vspace*{0mm}
	\centering
	\includegraphics[scale=1.3]{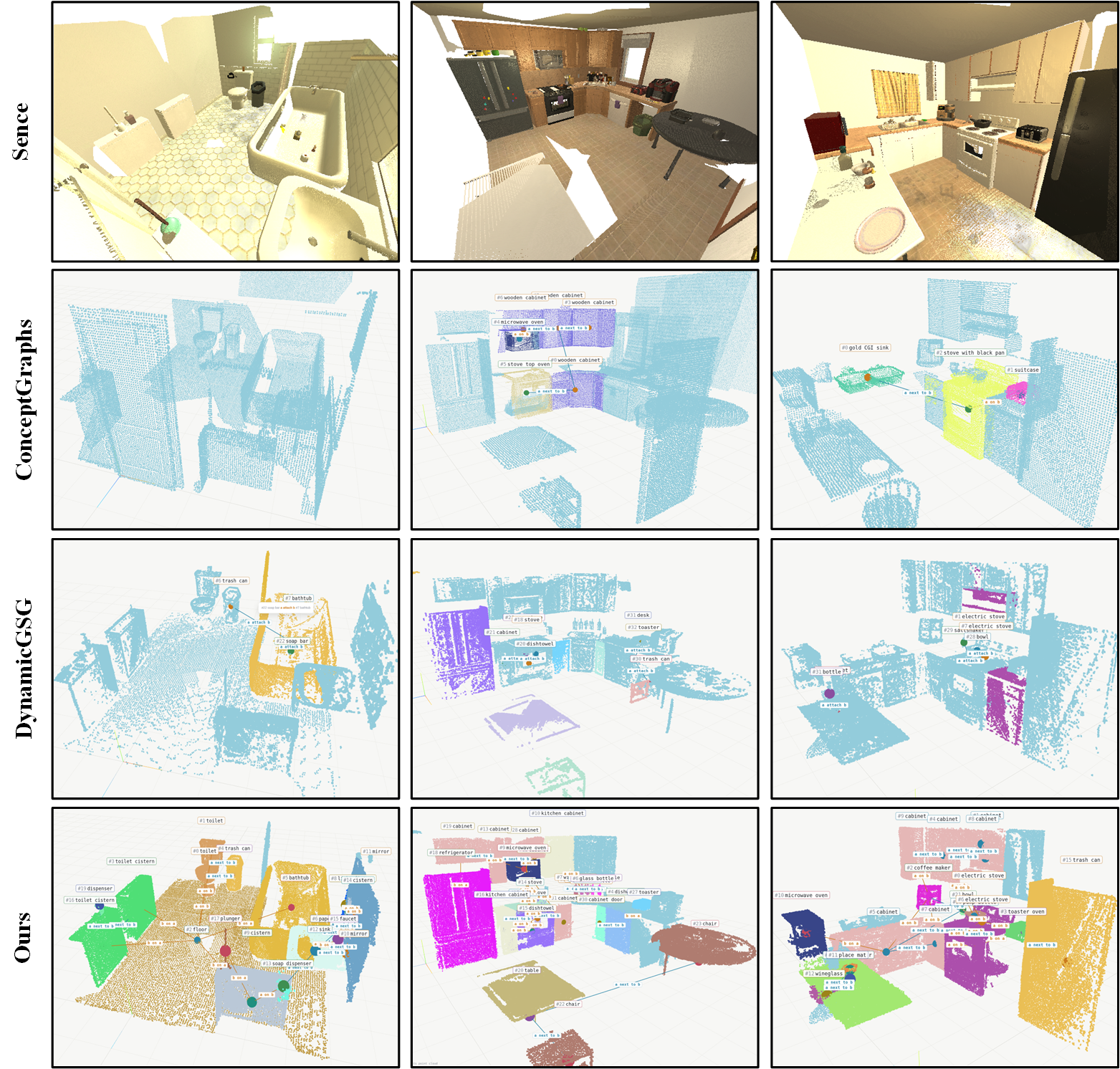}
	\vspace{-3mm}
	\caption{Qualitative comparison of spatial relationships on the Dyn-THOR dataset. Object point clouds are shown in blue by default. Objects that are correctly recognized and assigned spatial relationships are highlighted with colors corresponding to their semantic labels. The results demonstrate that the proposed method constructs accurate and semantically rich scene graphs.}
	\label{ai2thor_edges_visual}
	\vspace{-5mm}
\end{figure*}

\subsection{Edge-Level Evaluation}

We next evaluate the quality of scene graph edges, i.e., the spatial relationships between object nodes. The evaluation is conducted on the Dyn-THOR dataset by comparing the proposed method with ConceptGraphs~\cite{gu2024conceptgraphs} and DynamicGSG~\cite{ge2025dynamicgsg}.

Table~\ref{ai2thor_edges_table} shows that the proposed method predicts substantially more valid spatial relationships than the compared methods, producing an average of 15.5 edges per scene. This demonstrates that the constructed scene graph contains richer relational information while maintaining high-quality object nodes. The proposed method incorporates multi-granularity visual context together with geometric information for relationship reasoning, enabling more comprehensive prediction of spatial relationships. In contrast, DynamicGSG only establishes predefined hierarchical relations between different object categories rather than explicit spatial relationships, resulting in limited graph expressiveness. Although ConceptGraphs also performs candidate filtering using geometric proximity, its inability to update object nodes after object relocation and the lack of visual contextual information substantially reduce the number of valid relationships that can be inferred.

The proposed method also achieves the highest numbers of both coarse and fine matches, indicating superior accuracy in spatial relationship prediction. In particular, the coarse-level matching rate (coarse matches divided by evaluable edges) reaches approximately 43\%, compared with 22\% for DynamicGSG and only 8\% for ConceptGraphs. These results demonstrate that our method more reliably identifies whether a spatial relationship exists between neighboring objects. Furthermore, compared with the two baselines, the proposed method achieves the highest number of fine-grained matches, demonstrating its capability to infer not only the existence of a relationship but also its semantic type and direction.

\begin{table}[h]
\centering
\caption{Edge-level evaluation and ablation study on the Dyn-THOR dataset. PE: Predicted Edges, Eval.: Evaluable Edges, CM: Coarse Matches, FM: Fine Matches.}
\label{ai2thor_edges_table}
\begin{tabular*}{\columnwidth}{@{\extracolsep{\fill}}lcccc}
\toprule
Method & PE$\uparrow$ & Eval.$\uparrow$ & CM$\uparrow$ & FM$\uparrow$ \\
\midrule
ConceptGraphs~\cite{gu2024conceptgraphs} & 3.1 & 2.5 & 0.2 & 0.1 \\
DynamicGSG~\cite{ge2025dynamicgsg}       & 5.2 & 3.6 & 0.8 & 0.0 \\
\midrule
Ours w/o Visual Context           & 13.6 & \textbf{4.2} & 1.4 & \textbf{0.8} \\
Ours w/ Visual Context (Full)    & \textbf{15.5} & \textbf{4.2} & \textbf{1.8} & \textbf{0.8} \\
\bottomrule
\end{tabular*}
\end{table}

Furthermore, Table~\ref{ai2thor_edges_table} shows the contribution of visual context to spatial relation reasoning. The number of evaluable edges remains unchanged regardless of whether visual context is used, since candidate edge generation is governed solely by geometric proximity and is unaffected by this ablation. Incorporating visual context increases the number of predicted valid edges from 13.6 to 15.5 and improves coarse-grained relation accuracy from 1.4 to 1.8, while fine-grained matching accuracy remains unchanged at 0.8. This indicates that visual context primarily helps the model recognize spatial relationships that would otherwise be missed at the coarse-grained level when relying on geometric reasoning alone.

\subsection{Qualitative Analysis}

\textbf{Node visualization:} The qualitative results shown in Fig.~\ref{ai2thor_nodes_visual} further demonstrate the advantages of the proposed object difference detection strategy. On the Dyn-THOR dataset, ConceptGraphs cannot update object nodes after object relocation, leaving obsolete objects in the scene graph. DynamicGSG alleviates this problem through rendering-based comparison, but its reliance on renderings generated from unoptimized Gaussian primitives makes it sensitive to rendering artifacts, frequently causing missed detections or erroneous deletions. In contrast, the proposed method first optimizes Gaussian primitives before rendering and further verifies uncertain objects from an optimized observation viewpoint, resulting in substantially more accurate object reconstruction.

The superiority of the proposed method becomes more evident on the 3RScan dataset, as shown in Fig.~\ref{3rscan_nodes_visual}. Since the dataset is captured in real-world environments, motion blur, lighting variations, and imperfect camera trajectories often degrade rendering quality. Under these challenging conditions, DynamicGSG incorrectly removes many valid objects because the rendered appearance differs significantly from the observed image. Benefiting from optimized Gaussian representations and dual-view verification, our method preserves a more complete scene representation.

\begin{figure}[ht]
    \vspace*{3mm}
	\centering
	\includegraphics[scale=0.8]{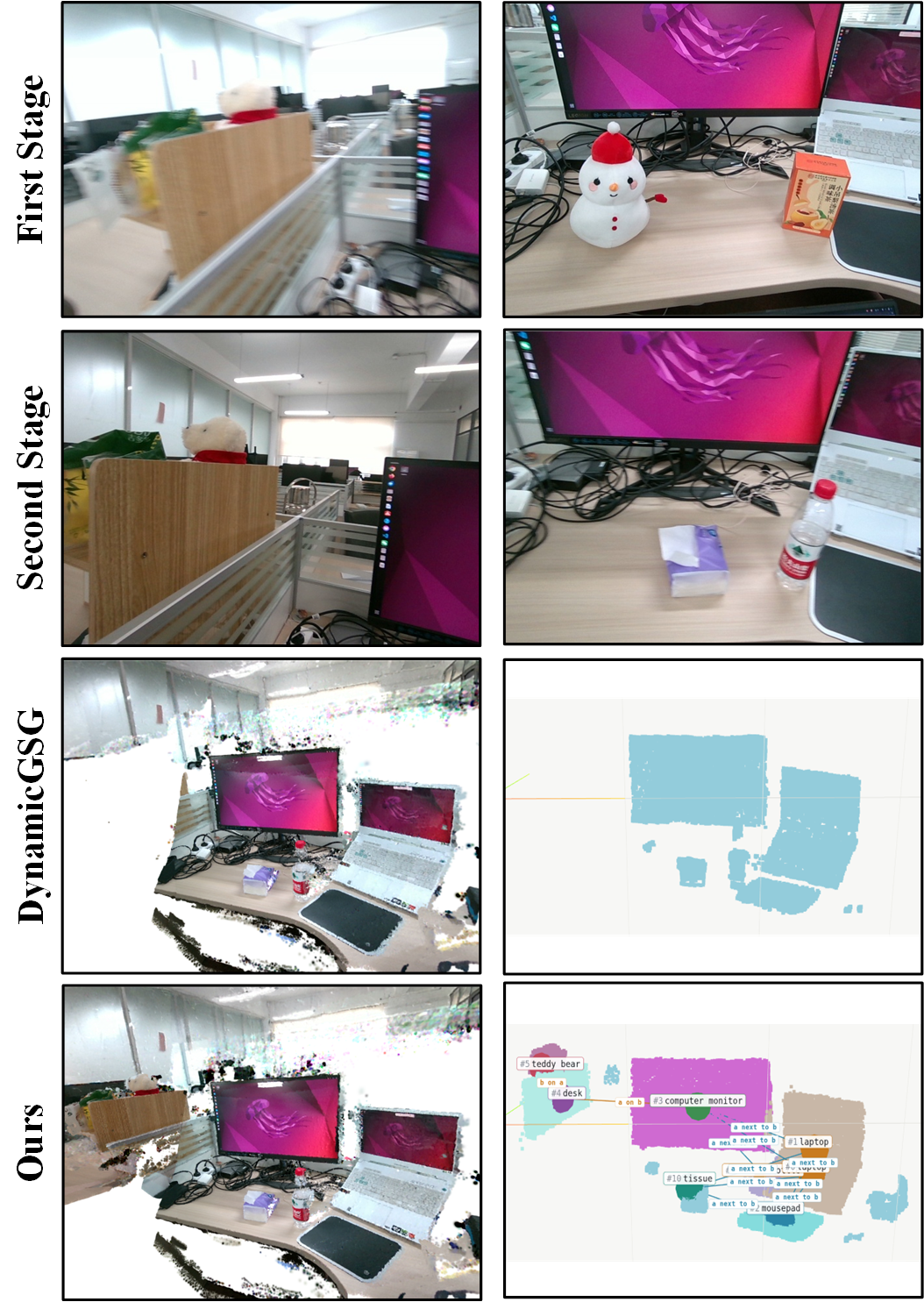}
	\vspace{-3mm}
	\caption{Qualitative comparison of scene graph construction in a real-world indoor environment. Compared with DynamicGSG, the proposed method achieves more accurate object updates and richer relationship reasoning, enabling the construction of a more complete and meaningful 3D scene graph.}
	\label{realsense_desk}
	\vspace{-5mm}
\end{figure}

\textbf{Edge visualization:} We compare the spatial relationships generated by different methods across multiple indoor environments on the Dyn-THOR dataset. As shown in Fig.~\ref{ai2thor_edges_visual}. From left to right are the ground truth, ConceptGraphs, DynamicGSG, and the proposed method. ConceptGraphs relies primarily on object labels and geometric information for large language model reasoning. Without explicit visual context, it predicts only a limited number of relationships, resulting in sparse scene graphs. DynamicGSG establishes only predefined hierarchical relations between object categories, making it difficult to represent the diverse spatial interactions that naturally occur in indoor environments. Furthermore, inaccurate object updates occasionally preserve obsolete or incorrectly reconstructed objects, which further degrades the quality of the inferred relationships.

In contrast, the proposed method combines accurate object updates with multi-granularity visual descriptions for spatial relationship reasoning. The visual context extracted from the object itself and its surrounding region provides complementary evidence beyond geometric information, helping the LLM identify spatial relationships that might otherwise be missed when relying on geometric cues alone. Consequently, the constructed scene graph contains richer relational information and provides a more complete representation of the spatial interactions in the scene.

We further evaluate the proposed method in real indoor environments. As shown in Fig.~\ref{realsense_desk}, our method consistently maintains accurate object representations even in scenes containing challenging materials, incomplete observations, and previously unseen object categories, demonstrating its generalization capability for dynamic scene graph construction.

\section{Conclusions}

This paper presented DSG, a dynamic 3D scene graph construction framework for indoor environments with changing object positions, combining dual-view rendering comparison for object update with visual-context-driven large language model reasoning for spatial relationship inference. Experimental results demonstrate that comparing optimized rather than raw Gaussian renderings substantially reduces false object deletions, while verification from an optimized observation viewpoint further identifies missing objects that cannot be reliably detected from the current viewpoint alone. Furthermore, incorporating visual context together with geometric information enables the LLM to identify a richer set of spatial relationships and improves coarse-grained relationship recognition. Extensive experiments on both simulated and real-world scenes validate the effectiveness of the proposed framework. In future work, we plan to extend DSG toward an end-to-end 4D dynamic scene graph framework for long-term scene understanding.

\bibliographystyle{IEEEtran}
\bibliography{ref}

@inproceedings{ge2025dynamicgsg,
  title={Dynamicgsg: Dynamic 3d gaussian scene graphs for environment adaptation},
  author={Ge, Luzhou and Zhu, Xiangyu and Yang, Zhuo and Li, Xuesong},
  booktitle={2025 IEEE/RSJ International Conference on Intelligent Robots and Systems (IROS)},
  pages={2232--2239},
  year={2025},
  organization={IEEE}
}

@inproceedings{lee2026embodiedsplat,
  title={EmbodiedSplat: Online Feed-Forward Semantic 3DGS for Open-Vocabulary 3D Scene Understanding},
  author={Lee, Seungjun and Wang, Zihan and Wang, Yunsong and Lee, Gim Hee},
  booktitle={Proceedings of the IEEE/CVF Conference on Computer Vision and Pattern Recognition},
  pages={23774--23784},
  year={2026}
}

@inproceedings{li2025hier,
  title={Hier-slam: Scaling-up semantics in slam with a hierarchically categorical gaussian splatting},
  author={Li, Boying and Cai, Zhixi and Li, Yuan-Fang and Reid, Ian and Rezatofighi, Hamid},
  booktitle={2025 IEEE International Conference on Robotics and Automation (ICRA)},
  pages={9748--9754},
  year={2025},
  organization={IEEE}
}

@article{li2025pg,
  title={Pg-slam: Photo-realistic and geometry-aware rgb-d slam in dynamic environments},
  author={Li, Haoang and Meng, Xiangqi and Zuo, Xingxing and Liu, Zhe and Wang, Hesheng and Cremers, Daniel},
  journal={IEEE Transactions on Robotics},
  year={2025},
  publisher={IEEE}
}

@inproceedings{zheng2025wildgs,
  title={Wildgs-slam: Monocular gaussian splatting slam in dynamic environments},
  author={Zheng, Jianhao and Zhu, Zihan and Bieri, Valentin and Pollefeys, Marc and Peng, Songyou and Armeni, Iro},
  booktitle={Proceedings of the IEEE/CVF Conference on Computer Vision and Pattern Recognition},
  pages={11461--11471},
  year={2025}
}

@inproceedings{gu2024conceptgraphs,
  title={Conceptgraphs: Open-vocabulary 3d scene graphs for perception and planning},
  author={Gu, Qiao and Kuwajerwala, Ali and Morin, Sacha and Jatavallabhula, Krishna Murthy and Sen, Bipasha and Agarwal, Aditya and Rivera, Corban and Paul, William and Ellis, Kirsty and Chellappa, Rama and others},
  booktitle={2024 IEEE International Conference on Robotics and Automation (ICRA)},
  pages={5021--5028},
  year={2024},
  organization={IEEE}
}

@inproceedings{wu2021scenegraphfusion,
  title={Scenegraphfusion: Incremental 3d scene graph prediction from rgb-d sequences},
  author={Wu, Shun-Cheng and Wald, Johanna and Tateno, Keisuke and Navab, Nassir and Tombari, Federico},
  booktitle={Proceedings of the IEEE/CVF Conference on Computer Vision and Pattern Recognition},
  pages={7515--7525},
  year={2021}
}

@article{chang2026rag,
  title={RAG-3DSG: Enhancing 3D Scene Graphs with Re-Shot Guided Retrieval-Augmented Generation},
  author={Chang, Yue and Chen, Rufeng and Zhang, Zhaofan and Chen, Yi and Tian, Yifan and Xie, Sihong},
  journal={arXiv preprint arXiv:2601.10168},
  year={2026}
}

@inproceedings{koch2024open3dsg,
  title={Open3dsg: Open-vocabulary 3d scene graphs from point clouds with queryable objects and open-set relationships},
  author={Koch, Sebastian and Vaskevicius, Narunas and Colosi, Mirco and Hermosilla, Pedro and Ropinski, Timo},
  booktitle={Proceedings of the IEEE/CVF Conference on Computer Vision and Pattern Recognition},
  pages={14183--14193},
  year={2024}
}

@article{kolve2017ai2,
  title={Ai2-thor: An interactive 3d environment for visual ai},
  author={Kolve, Eric and Mottaghi, Roozbeh and Han, Winson and VanderBilt, Eli and Weihs, Luca and Herrasti, Alvaro and Deitke, Matt and Ehsani, Kiana and Gordon, Daniel and Zhu, Yuke and others},
  journal={arXiv preprint arXiv:1712.05474},
  year={2017}
}

@inproceedings{liu2024grounding,
  title={Grounding dino: Marrying dino with grounded pre-training for open-set object detection},
  author={Liu, Shilong and Zeng, Zhaoyang and Ren, Tianhe and Li, Feng and Zhang, Hao and Yang, Jie and Jiang, Qing and Li, Chunyuan and Yang, Jianwei and Su, Hang and others},
  booktitle={European conference on computer vision},
  pages={38--55},
  year={2024},
  organization={Springer}
}

@inproceedings{kirillov2023segment,
  title={Segment anything},
  author={Kirillov, Alexander and Mintun, Eric and Ravi, Nikhila and Mao, Hanzi and Rolland, Chloe and Gustafson, Laura and Xiao, Tete and Whitehead, Spencer and Berg, Alexander C and Lo, Wan-Yen and others},
  booktitle={Proceedings of the IEEE/CVF international conference on computer vision},
  pages={4015--4026},
  year={2023}
}

@inproceedings{radford2021learning,
  title={Learning transferable visual models from natural language supervision},
  author={Radford, Alec and Kim, Jong Wook and Hallacy, Chris and Ramesh, Aditya and Goh, Gabriel and Agarwal, Sandhini and Sastry, Girish and Askell, Amanda and Mishkin, Pamela and Clark, Jack and others},
  booktitle={International conference on machine learning},
  pages={8748--8763},
  year={2021},
  organization={PmLR}
}

@article{rosinol2021kimera,
  title={Kimera: From SLAM to spatial perception with 3D dynamic scene graphs},
  author={Rosinol, Antoni and Violette, Andrew and Abate, Marcus and Hughes, Nathan and Chang, Yun and Shi, Jingnan and Gupta, Arjun and Carlone, Luca},
  journal={The International Journal of Robotics Research},
  volume={40},
  number={12-14},
  pages={1510--1546},
  year={2021},
  publisher={SAGE Publications Sage UK: London, England}
}

@article{hughes2022hydra,
  title={Hydra: A real-time spatial perception system for 3D scene graph construction and optimization},
  author={Hughes, Nathan and Chang, Yun and Carlone, Luca},
  journal={arXiv preprint arXiv:2201.13360},
  year={2022}
}

@inproceedings{keetha2024splatam,
  title={Splatam: Splat track \& map 3d gaussians for dense rgb-d slam},
  author={Keetha, Nikhil and Karhade, Jay and Jatavallabhula, Krishna Murthy and Yang, Gengshan and Scherer, Sebastian and Ramanan, Deva and Luiten, Jonathon},
  booktitle={Proceedings of the IEEE/CVF conference on computer vision and pattern recognition},
  pages={21357--21366},
  year={2024}
}

@inproceedings{hu2024cg,
  title={Cg-slam: Efficient dense rgb-d slam in a consistent uncertainty-aware 3d gaussian field},
  author={Hu, Jiarui and Chen, Xianhao and Feng, Boyin and Li, Guanglin and Yang, Liangjing and Bao, Hujun and Zhang, Guofeng and Cui, Zhaopeng},
  booktitle={European Conference on Computer Vision},
  pages={93--112},
  year={2024},
  organization={Springer}
}

@inproceedings{qin2024langsplat,
  title={Langsplat: 3d language gaussian splatting},
  author={Qin, Minghan and Li, Wanhua and Zhou, Jiawei and Wang, Haoqian and Pfister, Hanspeter},
  booktitle={Proceedings of the IEEE/CVF Conference on Computer Vision and Pattern Recognition},
  pages={20051--20060},
  year={2024}
}

@inproceedings{ye2024gaussian,
  title={Gaussian grouping: Segment and edit anything in 3d scenes},
  author={Ye, Mingqiao and Danelljan, Martin and Yu, Fisher and Ke, Lei},
  booktitle={European conference on computer vision},
  pages={162--179},
  year={2024},
  organization={Springer}
}

@article{rosinol20203d,
  title={3D dynamic scene graphs: Actionable spatial perception with places, objects, and humans},
  author={Rosinol, Antoni and Gupta, Arjun and Abate, Marcus and Shi, Jingnan and Carlone, Luca},
  journal={arXiv preprint arXiv:2002.06289},
  year={2020}
}

@article{kim20193,
  title={3-d scene graph: A sparse and semantic representation of physical environments for intelligent agents},
  author={Kim, Ue-Hwan and Park, Jin-Man and Song, Taek-Jin and Kim, Jong-Hwan},
  journal={IEEE transactions on cybernetics},
  volume={50},
  number={12},
  pages={4921--4933},
  year={2019},
  publisher={IEEE}
}

@article{bavle2023s,
  title={S-graphs+: Real-time localization and mapping leveraging hierarchical representations},
  author={Bavle, Hriday and Sanchez-Lopez, Jose Luis and Shaheer, Muhammad and Civera, Javier and Voos, Holger},
  journal={IEEE Robotics and Automation Letters},
  volume={8},
  number={8},
  pages={4927--4934},
  year={2023},
  publisher={IEEE}
}

@inproceedings{werby2024hierarchical,
  title={Hierarchical open-vocabulary 3d scene graphs for language-grounded robot navigation},
  author={Werby, Abdelrhman and Huang, Chenguang and B{\"u}chner, Martin and Valada, Abhinav and Burgard, Wolfram},
  booktitle={First Workshop on Vision-Language Models for Navigation and Manipulation at ICRA 2024},
  year={2024}
}

@article{werby2025keysg,
  title={KeySG: Hierarchical Keyframe-Based 3D Scene Graphs},
  author={Werby, Abdelrhman and Rotondi, Dennis and Scaparro, Fabio and Arras, Kai O},
  journal={arXiv preprint arXiv:2510.01049},
  year={2025}
}

@article{schmid2024khronos,
  title={Khronos: A unified approach for spatio-temporal metric-semantic slam in dynamic environments},
  author={Schmid, Lukas and Abate, Marcus and Chang, Yun and Carlone, Luca},
  journal={arXiv preprint arXiv:2402.13817},
  year={2024}
}

@inproceedings{wald2019rio,
  title={Rio: 3d object instance re-localization in changing indoor environments},
  author={Wald, Johanna and Avetisyan, Armen and Navab, Nassir and Tombari, Federico and Nie{\ss}ner, Matthias},
  booktitle={Proceedings of the IEEE/CVF International Conference on Computer Vision},
  pages={7658--7667},
  year={2019}
}

@inproceedings{lian2025describe,
  title={Describe anything: Detailed localized image and video captioning},
  author={Lian, Long and Ding, Yifan and Ge, Yunhao and Liu, Sifei and Mao, Hanzi and Li, Boyi and Pavone, Marco and Liu, Ming-Yu and Darrell, Trevor and Yala, Adam and others},
  booktitle={Proceedings of the IEEE/CVF International Conference on Computer Vision},
  pages={21766--21777},
  year={2025}
}

@article{qin2025general,
  title={A General Optimisation-Based Framework for Global Pose Estimation With Multiple Sensors},
  author={Qin, Tong and Cao, Shaozu and Pan, Jie and Shen, Shaojie},
  journal={IET Cyber-Systems and Robotics},
  volume={7},
  number={1},
  pages={e70023},
  year={2025},
  publisher={Wiley Online Library}
}

@article{lv2023multimodality,
  title={A multimodality scene graph generation approach for robust human--robot collaborative assembly visual relationship representation},
  author={Lv, Jianhao and Zhang, Rong and Li, Xinyu and Liu, Shimin and Liu, Tianyuan and Zhang, Qi and Bao, Jinsong},
  journal={IEEE Transactions on Industrial Informatics},
  volume={20},
  number={3},
  pages={3242--3251},
  year={2023},
  publisher={IEEE}
}

@article{ye2025csgrasp,
  title={CSGrasp: Category-level semantically-aware grasping method},
  author={Ye, Chao and Yao, Yangdong and Lin, Weiyang and Yang, Xuebo and Qiu, Jianbin},
  journal={IEEE Transactions on Industrial Electronics},
  year={2025},
  publisher={IEEE}
}

@article{li2025dqo,
  title={DQO-MAP: Real-Time Object-Level SLAM with Dual Quadrics and Gaussians},
  author={Li, Haoyuan and Ye, Ziqin and Hao, Yue and Lin, Weiyang and Ye, Chao},
  journal={IEEE Robotics and Automation Letters},
  volume={11},
  number={2},
  pages={1034--1041},
  year={2025},
  publisher={IEEE}
}

@article{zhang2026long,
  title={Long-Term Dynamic Object Relocalization for Mobile Robots in Human--Robot Coexisting Environments},
  author={Zhang, Ying and Bi, Wenfu and Yin, Maoliang and Qu, Hongqiang and Zhang, Cui-Hua and Hua, Changchun and Wen, Guilin},
  journal={IEEE Transactions on Industrial Informatics},
  year={2026},
  publisher={IEEE}
}

@article{qi2023instance,
  title={Instance-incremental scene graph generation from real-world point clouds via normalizing flows},
  author={Qi, Chao and Yin, Jianqin and Xu, Jinghang and Ding, Pengxiang},
  journal={IEEE Transactions on Circuits and Systems for Video Technology},
  volume={34},
  number={2},
  pages={1057--1069},
  year={2023},
  publisher={IEEE}
}

@article{feng2025history,
  title={History-enhanced 3D scene graph reasoning from RGB-D sequences},
  author={Feng, Mingtao and Yan, Chenbo and Wu, Zijie and Dong, Weisheng and Wang, Yaonan and Mian, Ajmal},
  journal={IEEE Transactions on Circuits and Systems for Video Technology},
  volume={35},
  number={8},
  pages={7667--7682},
  year={2025},
  publisher={IEEE}
}

@article{hou2026spatial,
  title={Spatial Multimodal Knowledge Driven 3D Scene Graph Prediction with Vision-Language Model},
  author={Hou, Haoran and Feng, Mingtao and Wu, Zijie and Guo, Yulan and Wang, Yaonan and Mian, Ajmal},
  journal={IEEE Transactions on Circuits and Systems for Video Technology},
  year={2026},
  publisher={IEEE}
}

\end{document}